\pdfoutput=1

\documentclass[11pt]{article}
\usepackage[export]{adjustbox}
\usepackage[final]{acl}
\usepackage{graphicx}
\usepackage{times}
\usepackage{latexsym}
\usepackage{multirow}

\usepackage[T1]{fontenc}

\usepackage[utf8]{inputenc}
\usepackage{wrapfig}
\usepackage{microtype}                         
\usepackage{booktabs}
\usepackage{inconsolata}
\usepackage{amsmath}
\usepackage{amssymb}

\usepackage{rotating}
\usepackage{times}
\usepackage{latexsym}
\usepackage{makecell}

\usepackage[T1]{fontenc}

\usepackage[utf8]{inputenc}
\usepackage{comment}
\usepackage{microtype}

\usepackage{inconsolata}

\usepackage{graphicx}
\usepackage{subfigure}
\usepackage{color}
\usepackage{soul}
\usepackage{colortbl}
\usepackage{xcolor}

\usepackage{pifont}       
\usepackage{bbding}       
\usepackage{fontawesome}  

\title{CausalMACE\includegraphics[scale=1,valign=c]{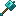}: Causality Empowered Multi-Agents in Minecraft Cooperative Tasks}

\author{Qi Chai$^{1}$, Zheng Zhang$^{1}$, Junlong Ren$^{1}$,
        Deheng Ye$^{2}$, Zichuan Lin$^{2}$, Hao Wang$^{1}$\thanks{Corresponding Author.} \\
        $^1$The Hong Kong University of Science and Technology (Guangzhou),
        $^2$Tencent \\
        \texttt{\{qchai315, zzhang302, jren686\}@connect.hkust-gz.edu.cn}\\ 
        \texttt{\{dericye, zichuanlin\}@tencent.com}, 
        \texttt{haowang@hkust-gz.edu.cn}}

\begin{document}
\maketitle

\begin{abstract}
Minecraft, as an open-world virtual interactive environment, has become a prominent platform for research on agent decision-making and execution.
Existing works primarily adopt a single Large Language Model (LLM) agent to complete various in-game tasks. 
However, for complex tasks requiring lengthy sequences of actions, single-agent approaches often face challenges related to inefficiency and limited fault tolerance. Despite these issues, research on multi-agent collaboration remains scarce. 
In this paper, we propose CausalMACE, a holistic causality planning framework designed to enhance multi-agent systems, in which we incorporate causality to manage dependencies among subtasks. 
Technically, our proposed framework introduces two modules: an overarching task graph for global task planning and a causality-based module for dependency management, where inherent rules are adopted to perform causal intervention. Experimental results demonstrate our approach achieves state-of-the-art performance in multi-agent cooperative tasks of Minecraft. 

\end{abstract}

\section{Introduction}

In recent years, Large Language Models (LLMs) have shown significant abilities in various domains~\cite{xu2023symbol,wang2023can,cavar2024computing,zubiaga2024natural}. Beyond these fundamental domains, interest has increased in how to utilize LLMs' capabilities to make decisions in an open-world environment.

Minecraft, as a virtual interactive environment, offers a unique platform for such research. Various approaches ~\cite{wang2024describe,wangvoyager,wang2024jarvis,lioptimus} have been developed within Minecraft to explore the decision-making capabilities of LLMs, achieving significant progress on in-game tasks. However, complex tasks that require lengthy sequences of action or collaboration remain a challenge, since single-agent approaches often face issues of inefficiency and limited fault tolerance.

To fully unlock the potential of LLMs for decision-making in open-world environments, it is essential to explore the usage of multi-agent systems. Research has shown that LLM agent teams can outperform a single agent in writing~\cite{chanchateval}, reasoning~\cite{xu2024magic} and code generation~\cite{hongmetagpt}. However, existing methods often emphasize role assignment or communication to enhance performance, which does not fully exploit the parallelization potential of multi-agent systems, thus limiting their application in open-world decision-making tasks.

\begin{figure}[!tbp]
    
    \centering
    \subfigure[Initial State.]{\raisebox{0\columnwidth}{\includegraphics[width=0.48\linewidth]{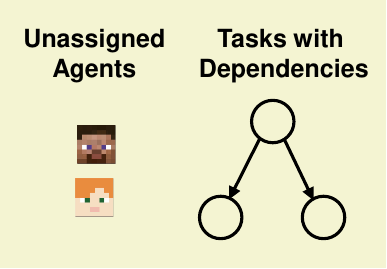}\label{fig:1a}\hspace{0.01\linewidth}}}
    \subfigure[Ignoring Dependencies.]{\hspace{0.01\linewidth}{\includegraphics[width=0.48\linewidth]{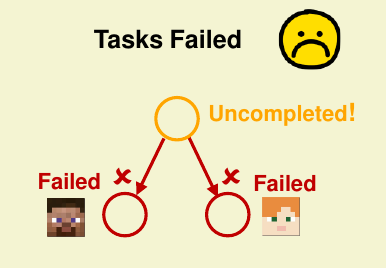}\label{fig:1b}}}
    \subfigure[Following Dependencies (Ours).]{\raisebox{0\columnwidth}{\includegraphics[width=1\linewidth]{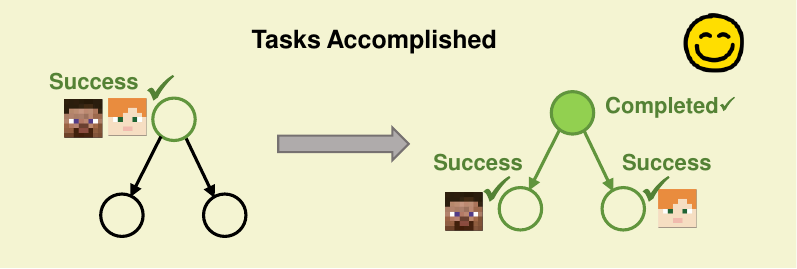}\label{fig:1c}}}
    \vspace{-3mm}
    \caption{\textbf{The impact of dependencies among tasks.} (a) describes tasks with dependency (arrows). (b) shows the consequences of ignoring dependencies: agents failed to execute due to unmet prerequisites. (c) demonstrates 
the proper dependency processing: sequentially complete root node tasks before leaf nodes.}
    \label{fig:1}
    \vspace{-1.5em}
\end{figure}

While existing methods~\cite{chen2023agentverse, dong2024villageragent} attempt to enhance the performance of multi-agent systems, they often encounter bottlenecks stemming from several unaddressed issues. The first issue is that current multi-agent methods in Minecraft lack global task planning. While local task planning is relatively flexible, the absence of global planning may cause agents to deviate from the original plan after multiple iterations gradually. Another issue is that existing methods fail to consider the dependencies that exist within subtasks. This can result in assigned tasks being unachievable thereby reducing efficiency, as shown in Figure \ref{fig:1}. 

Given that the organization of subtasks in open-world tasks is inherently shaped by the environment's rules, we propose that causal relations should exist between these rules and the dependencies among subtasks. When these causal relations are clearly identified, subtasks can naturally be structured into a cohesive task graph. 
This assumption enables us to use causality~\cite{yuan2023instrumental, wu2024decot} to guide tasks and to construct a coherent global task graph. Based on this premise, we propose CausalMACE (\textbf{\underline{Causal}}ity Empowered \textbf{\underline{M}}ulti-\textbf{\underline{A}}gents in Minecraft \textbf{\underline{C}}ooperativ\textbf{\underline{E}} Tasks), 
a global causality planning framework. Specifically, we incorporate two new modules: one for maintaining an overarching task graph for global task planning, and another that utilizes causality to construct and manage the dependencies among subtasks. CausalMACE achieves an average performance improvement of 12\% in multi-agent cooperative tasks and 7\% in single-agent tasks, achieving state-of-the-art results. 

Our contributions can be concluded as follows:

\begin{itemize}
\setlength{\itemsep}{0pt}
\setlength{\parsep}{0pt}
\setlength{\parskip}{0pt}
    \item We propose a novel framework for multi-agent cooperative tasks of Minecraft. An overarching task graph is proposed for global planning that facilitates comprehensive tasks.
    \item We propose to leverage causality to manage and build dependencies between subtasks. This ensures the task graph is aligned with the inherent rules of open-world environments.
    \item Experimental results demonstrate our framework surpasses existing baselines in multi-agent cooperative tasks and also achieves competitive results in single-agent tasks.
\end{itemize}

\begin{figure*}
  \centering 
  \includegraphics[width=\textwidth]{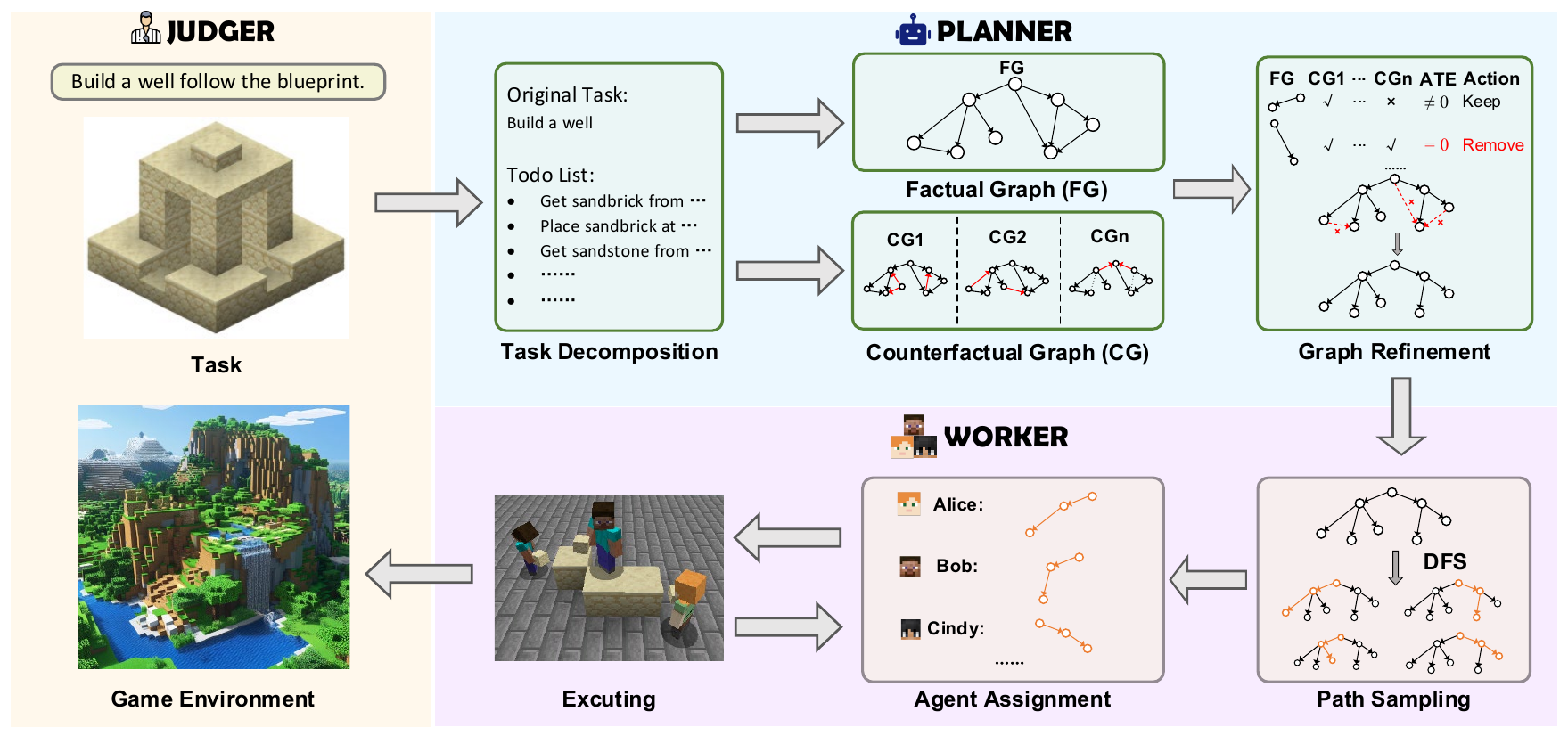} 
  \caption{\textbf{Overview of the framework.} Our framework contains Judger, Planner and Worker. The \textbf{Judger} defines the objective and gives the environment feedback. The \textbf{Planner} decomposes the task into subtasks and constructs a dependency graph via game rules. The \textbf{Worker} assigns subtasks to agents to process. \textit{ATE} denotes the average treatment effect, which is introduced in Section \ref{sec:Planner}.} 
  \label{fig:framework} 
  \vspace{-1.5em}
\end{figure*}

\section{Related Work}
\subsection{Causal Discovery}
The causal discovery method can be delineated into two groups, traditional approaches and large language models (LLMs) based methods. Traditional causal discovery approaches rely mainly on statistical techniques and algorithms to uncover causal structures in data. Several methods ~\cite{xiang2013lasso,chickering2015selective,ramsey2015scaling} calculate scores to investigate the entire range of possible edges, with the aim of finding a graph that fits the data the best. Some other researches~\cite{spirtes2001causation,spirtes2013causal} use conditional independence tests to infer causal structures, identifying dependencies among variables.

Recent advances in LLMs have offered the possibility of utilizing LLMs in causal discovery. Some studies attempt to use LLMs as an alternative tool for conditional independence tests in the traditional causal discovery process~\cite{cohrs2024large} or build causal graphs beyond the Markov equivalent class~\cite{long2023causal}. Despite these, there are also several data-free methods~\cite{kiciman2023causal,zevceviccausal,zhang2024causal}, which show that LLMs can identify causal structures by interpreting metadata and natural language, similar to how human experts apply domain knowledge to construct causal models.

\subsection{Multi-agent Systems}
Due to the growing potential of multi-agent systems in addressing complex problems, research on this topic has been expanding rapidly. With the integration of LLMs, multi-agent collaboration has shown significant promise across various tasks, enhancing individual agent performance. Existing approaches can be categorized into two strategies: role assignment and communication management.

For role assignment, CAMEL ~\cite{li2023camel} assigns roles to agents to optimize collaborative behaviors and mitigate hallucinations. MetaGPT ~\cite{hongmetagpt} leverages role specialization to coordinate multiple LLMs in simulating real-world software development workflows. Triad ~\cite{zong2024triad} addresses knowledge base question answering tasks by assigning three distinct roles, generalist, decision maker, and advisor, to agents based on LLMs. For communication management, ChatEval ~\cite{chanchateval} organizes multiple agents to debate and express their opinions, effectively reducing biases in text evaluation. DyLAN ~\cite{liu2024dynamic} improves collaboration efficiency by dynamically managing the composition of agent teams. Magic ~\cite{xu2024magic} employs probabilistic graphical models to enhance reasoning and facilitate interaction within agent teams. However, these methods predominantly operate in static temporal frameworks, neglecting the dynamic temporal coordination crucial for fully exploiting the parallelization potential inherent in multi-agent systems.

\subsection{Agents in Minecraft} 
Minecraft, which offers a rich environment for agent research, has emerged as a versatile platform for the testing of intelligent agents. Several researches~\cite{zhou2024learning,yuan2024pre,li2024auto} focus on low-level control policies with reinforcement learning (RL). With the development of LLMs, researchers attempt to enhance agents' capabilities by utilizing LLMs to process complex environmental feedback. DEPS~\cite{wang2024describe} pioneers this transition by implementing the first LLM-powered agent framework in Minecraft. Voyager~\cite{wangvoyager} adopts skill libraries to boost task execution efficiency. Additionally, techniques like multi-modal memories and knowledge graphs have been employed to enhance agents' adaptability and task performance~\cite{wang2024jarvis,lioptimus}.

Building on the progress of single-agent studies, some researchers have begun exploring the potential of multi-agent collaboration within Minecraft. Multi-agent systems offer significant advantages in task allocation and resource management, enabling more effective handling of complex scenarios in the game. AgentVerse~\cite{chen2023agentverse}, which is the initial study, has started to address challenges related to communication and coordination. It divides its framework into four components, each providing specific guidance to multiple LLM agents, effectively organizing agent groups, and outperforming single-agent systems. VillagerAgents~\cite{dong2024villageragent} attempts to break down the task into subtasks that can be processed in parallel. However, these approaches often overlook the causal relations between the game rules and dependencies among subtasks during task decomposition, limiting cooperative efficiency in complex tasks.

\section{Method}

\begin{figure*}[t]
  \centering
  \subfigure[]{\raisebox{-0.006\columnwidth}{\includegraphics[width=0.22\linewidth]{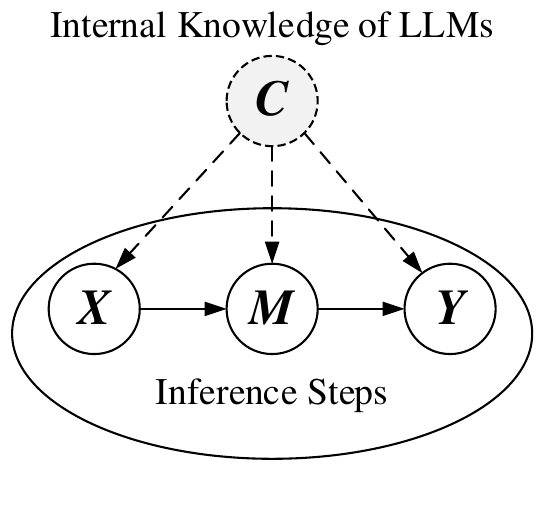}\label{fig:2a}\hspace{0.03\linewidth}}}
  \subfigure[]{\raisebox{0\columnwidth}{\includegraphics[width=0.22\linewidth]{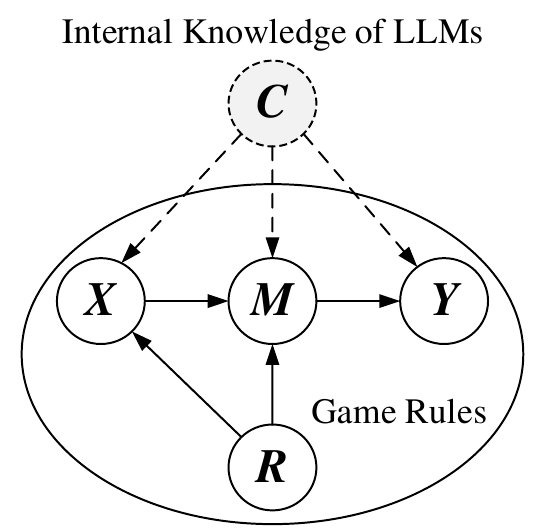}\label{fig:2b}\hspace{0.03\linewidth}}}
  \subfigure[]{\includegraphics[width=0.22\linewidth]{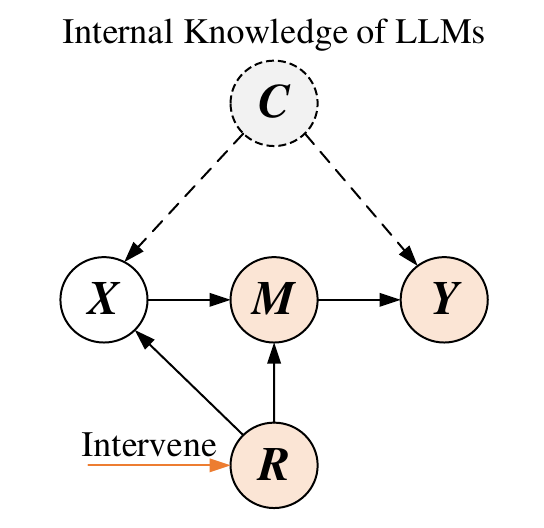}\label{fig:2c}\hspace{0.03\linewidth}}
  \subfigure[]{\includegraphics[width=0.22\linewidth]{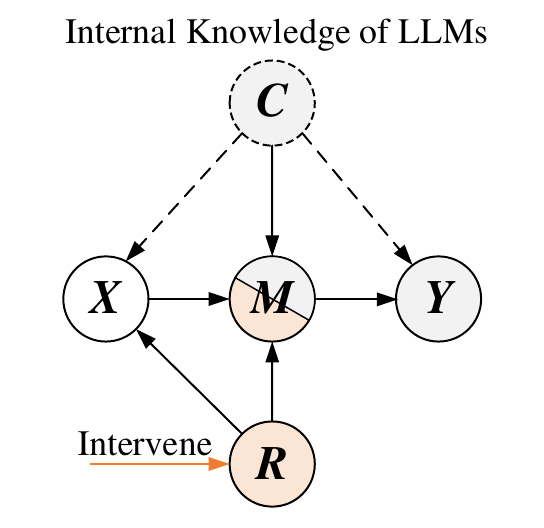}\label{fig:2d}}
  \vspace{-0.5em}
  \caption {\textbf{Procedures for generating the dependencies.} $X$ and $Y$ denote the input and output, $M$ is the inference steps of LLMs and $C$ represents the internal knowledge of LLMs.  (a) and (b) indicate the procedures with and without game rules $R$, respectively, where the dotted arrow denotes the uncertain causal edges. (c) shows the successful intervention procedure, (d) reveals the failure cases where the effect of $R$ on $M$ is blocked by $C$.}
  \vspace{-0.5em}
\end{figure*}

\subsection{Preliminary}
\noindent\textbf{Causality.} It represents the causal-and-effect relationships between variables. Causality seeks to understand how changes in one variable $X$ (the cause) \textit{directly affect} another variable $Y$ (the effect), i.e., $X \rightarrow Y$.
The Structural Causal Model (SCM)~\cite{pearl2009causal} provides a formal framework for identifying and leveraging causality. It employs directed acyclic graphs (DAGs) to represent causal structures, where nodes represent distinct variables and 
edges denote causal relationships between them.

\noindent\textbf{Mediator.} $M$ is the variable between $X$ and $Y$, where $X \rightarrow M$ and $M \rightarrow Y$. In this situation, the causal effect of $X$ can reach $Y$ by $X \rightarrow M \rightarrow Y$. Such a chain can be called a \textbf{causal path} so long as the causal effect can be transmitted.

\noindent\textbf{Confounder.} $C$ is a variable that affects both $X$ and $Y$, i.e., $C \rightarrow X$ and $C \rightarrow Y$. It may create a spurious association between $X$ and $Y$.

\noindent\textbf{Instrumental variable.} Given a causal path $X \rightarrow Y$, an instrumental variable is a variable that satisfies the following conditions: 1) It directly affects $X$, 2) No path exists from the instrumental variable to any confounder $C$ that affects both $X$ and $Y$, 3) It has no direct effect on $Y$.
\subsection{Overview}
Our proposed framework is shown in Figure \ref{fig:framework}, which contains three main components: Judger, Planner, and Worker. The Judger initiates tasks within the game. It defines the objectives the agents need to achieve and provides feedback and evaluations of their performance. 

Upon receiving a task from the Judger, the Planner decomposes it into a series of subtasks. It identifies the dependencies between these subtasks by leveraging the game rules and refines the resulting task graph using causal inference. This refined graph, which represents the structured dependencies and execution orders, is then passed on to the Worker for further action.

Technically, the Worker processes the refined task graph provided by the Planner. It employs a depth-first search (DFS) algorithm to explore all possible execution paths within the graph and then assigns these tasks to the agents for execution. The agents operate autonomously with reflection. Their actions are processed by the Judger, which also assesses the task execution.

\subsection{Judger Interface}

The Judger serves as an interface component that mediates interactions between agents and the game environment. It functions as both an action validator and a performance evaluator within the framework. When agents attempt to execute actions, the Judger processes these actions through the game environment and returns updated state information. Additionally, it incorporates evaluation functions that compute various performance metrics based on agent behaviors and outcomes. 
The Judger maintains a clear separation between the game environment and the agents' actions while providing standardized interfaces for both action processing and performance assessment.

\subsection{Planner with Casual Intervention}
\label{sec:Planner}

The Planner aims to transform a given task into a structured set of subtasks, which is represented by a causal dependency graph. 
The entire procedure involves three steps: task decomposition, initial graph construction, and graph refinement.

We enhance the global task planning capability. Upon receiving an input task from the Judger, the Planner employs LLMs to directly decompose it into a set of subtasks, $S = \{s_1, s_2, \dots, s_n\}$. These subtasks preserve global goal consistency with the input task. Then, we use the set of game rules  $R = \{r_1, r_2, \dots, r_k\}$ as explicit guidelines to identify direct dependencies between subtasks. These rules give an initial version of the global task dependency graph $G_{\text{init}}$ and ensure that it is structured logically.
However, it is observed $G_{\text{init}}$ may contain extra or incorrect edges due to hallucinations and biases stemming from the LLMs' internal knowledge \cite{yuan2023revisiting,wu2024decot}.

To resolve this issue, we aim to find out the actual dependencies between subtasks and refine $G_{\text{init}}$. 
Specifically, we define two subtasks, $s_p$ and $s_q$, as \textit{input} $X$, and their dependencies given by LLMs as \textit{output} $Y$. The \textit{mediator} that maps $X$ to $Y$ is denoted as $M$, which refers to the LLM inference process.
Since LLMs are used as the resolver, their internal knowledge may have a potential influence on $X$, $Y$ and $M$. Therefore, we define the internal knowledge of LLMs as the \textit{confounder} $C$, as illustrated in Figure \ref{fig:2a}.
Then, we introduce the set of game rules $R$ as external knowledge. As the dependencies between subtasks should be constrained by the rules, $R$ has a causal effect on both \textit{input} $X$ and \textit{mediator} $M$. This process is shown in Figure \ref{fig:2b}.

Next, we aim to ensure that the final \textit{output} $Y$ adheres to the set of game rules $R$. Since $M$ is the only controllable variable that directly affects $Y$, it is necessary to discern whether the primary effect of the inference steps $M$ comes from $R$ or internal knowledge $C$.
As tracking the causal relationship between $C$ and $M$ is difficult, an alternative method is to leverage the variable $R$ as an instrumental variable~\cite{yuan2023instrumental} instead, which can reach $Y$ through the causal path $R \rightarrow M \rightarrow Y$.

By intervening on $R$, we can observe how it affects $Y$. This effect is the average treatment effect (ATE), which allows us to estimate the causal effect on $Y$.
Specifically, for each game rule $r_i$, we query LLM for a counterfactual rule $r_i^*$, which is contradictory to $r_i$. For example, the counterfactual rule for ``\textit{You must have a block before you place it.}'' can be expressed as: ``\textit{You can place a block even if you do not have it}''. This procedure generates a modified set of rules $R_i^* = \{r_1,r_2,...,r_i^*,...,r_k\}$ where only $r_i$ is replaced while other rules remain unchanged. Following the modified set of rules as explicit guidelines, while taking $s_p$ and $s_q$ as \textit{input} $X$, we observe whether the \textit{output} $Y$ changes. Therefore, for the given $X$ and its inference steps $M$, the ATE of the certain rule $r_i$ is defined as:
\begin{equation}
  \label{eq:1}
  \begin{aligned}
      ATE(r_i, X) = \mathbb{E}(Y | X, M, do(R)) - \\
             \mathbb{E}(Y | X, M, do(R_i^*)),
  \end{aligned}
\end{equation}
where $do(\cdot)$ denotes the causal intervention operator ~\cite{pearl2009causal}, representing active manipulation of the variable, forcibly setting it to a given state.

This equation aligns with intuition, as if $r_i$ does not affect the output $Y$ (i.e., $ATE(r_i, X)=0$), which means if $r_i$ is replaced with $r_i^*$, the output $Y$ will remain consistent with the previous. 

To aggregate the causal effects across all individual rules into a unified impact, we compute the expectation over all $ATE(r_i, X)$ by averaging the individual effects across all rules:
\begin{equation}
  \label{eq:4}
  \begin{aligned}
  ATE(R, X) &= \mathbb{E}_i(ATE(r_i, X)) \\
            &= \frac{1}{k}\Sigma_{i=0}^{k}ATE(r_i,X).
  \end{aligned}
\end{equation}

With the calculated ATE, we proceed to refine the initial graph $G_{init}$.
For each edge in $G_{init}$, if it is correctly generated by the game rule set $R$ with given input node pair $X_i$, then $ATE(R, X_i) \neq 0$. This indicates that the set of game rules has a measurable influence on the outcome, as illustrated in Figure \ref{fig:2c}. When $ATE(R, X_i) = 0$, it implies that the edge is independent of the entire set of game rules, as demonstrated in Figure \ref{fig:2d}.
Therefore, such edges should be removed to ensure that all edges adhere to the set of game rules.

\subsection{Worker with Agent Assignment}
The Worker is responsible for translating the dependency graph of subtasks into concrete action plans for each agent, ensuring that the subtasks are executed efficiently and in alignment with the dependencies. The Worker takes the dependency graph given by the Planner as input and each agent will output the specific actions. The process begins with analyzing the dependency graph with the DFS algorithm.
This step identifies all the possible paths comprising the required subtasks while respecting their dependencies. Each path represents a sequence of subtasks that need to be executed. To accomplish the whole task, agents need to finish all the paths. Once an agent finishes its assigned path, it will be assigned to another.

To balance work distribution, the Worker needs to track how “busy” each task path is. Therefore, we introduce a new variable for each path: the busy rate $br$. This variable simultaneously captures two factors, assigned agent numbers in each path and agent density near the path entries. Consider a scenario where $k$ agents are executing path $p$, the busy rate $br$ of the path $p$ is computed as:
\begin{equation}
  \label{eq:5}
  br_p = \Sigma_{i=0}^k \frac{1}{d_i}, 
\end{equation}
where $d_i$ represents the number of subtasks between the current task of agent $i$ and the path entrance. The busy rate of a path $p$ dynamically increases with the number of agents newly assigned to it. It reflects the involvement of the current path and serves as an indicator of potential resource contention. While assigning an agent to a new path, the Worker prioritizes the path $p^*$ with the minimum busy rate, i.e., 
\begin{equation}
  \label{eq:6}
  p^* = \mathop{\arg\min}\limits_{p} {(br_p)}.
\end{equation}

\begin{table*}[t!]
\centering
\begin{adjustbox}{width=\linewidth}
\footnotesize
\begin{tabular}{lcccccccccc}
\toprule
\multirow{3}{*}{\textbf{Method}} &  \multirow{3}{*}{\textbf{\makecell{Agent\\Number\\\includegraphics[scale=0.05,valign=c]{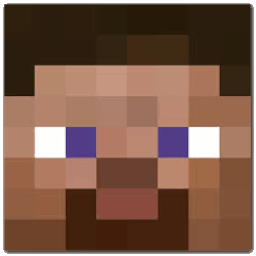}}}} & \multicolumn{3}{c}{\textbf{Construction Avg. Score}} & \multicolumn{2}{c}{\textbf{Escape Avg. Score}}&  \multicolumn{3}{c}{\textbf{Cooking Avg. Score}} & \textbf{All Avg.$^*$ Score}\\ 
\cmidrule(lr){3-5} \cmidrule(lr){6-7} \cmidrule(lr){8-10} \cmidrule(lr){11-11}
~ & ~ & \textbf{CR} & \textbf{VHR } &  \textbf{Efficiency}  &   \textbf{CR} &  \textbf{Efficiency}  &  \textbf{CR} & \textbf{ACR} &  \textbf{Efficiency} & \textbf{CR}     \\
~ & ~ & (\%)& (\%)& (\%/min)& (\%)& (\%/min)& (\%)&(\%)&(\%/min) &(\%) \\
\midrule
AgentVerse (ICLR'23)  & 2 & - & - & - & - & - & 29.75 & 48.64 & 3.54 & 29.75 \\ 
\midrule
\multirow{2}{*}{{VillagerAgent (ACL'24)}} & 2 & 36.45 & 49.05 & 3.88 & 73.29 & 149.4 & 73.75 & 58.11 & 6.98 & 56.90 \\ 
 & 3 & 52.17 & 61.02 & 6.26 & 69.78 & 227.4 & 85.26 & 55.60 & \textbf{21.90} & 68.82\\  
\midrule
\multirow{3}{*}{{CausalMACE (Ours)}} & 2 & 56.59 & 63.17 & 8.94 & \textbf{77.08} & 246.71 & 79.10 & 65.53 & 7.17 & 68.76 \\
 & 3 & 65.45 & 69.30 & \textbf{9.58} & 72.28 & \textbf{276.67} & 86.00 & 70.12 & 13.92 & 75.38\\
 \rowcolor{black!15} \cellcolor{white}& \cellcolor{white}6 & \textbf{76.04} & \textbf{78.99} & 8.20 & 69.25 & 169.59 & \textbf{88.75} & \textbf{72.30} & 16.31 & \textbf{81.09}\\
\bottomrule
\end{tabular}
\end{adjustbox}
\caption{\textbf{Comparison on cooperative tasks.} \textit{CR} denotes completion rate, \textit{VHR} denotes view hit rate, \textit{ACR} denotes agent contribution rate and \textit{Avg.$^*$} denotes the weighted average based on the number of tasks. With the same number of agents, our method shows improvements across all settings. 
}
\label{Main Result}
\vspace{-0.75em}
\end{table*}

Once an agent receives its designated path, it begins executing each subtask in the path sequentially. During execution, the agent interacts with the environment, completing actions as specified by the subtasks. Following the ReAct framework~\cite{yaoreact}, the agent iteratively generates actions and observes the environment until it makes a successful movement in the environment. The result of movement and environmental feedback will be recorded and the agent will first decide the statement of its current subtask and then update its strategy through a self-reflection process~\cite{ji2023towards}. This procedure will last until all paths are finished or the Judger returns an end signal.

\section{Experiment}

\subsection{Experiment Setup}
\subsubsection{Environment} 
We set up Minecraft Java Edition Server version 1.19.2 and host it on Ubuntu 20.04. Following \citet{wangvoyager,dong2024villageragent}, Mineflayer~\cite{mineflayer}, a JavaScript mod is installed to establish a platform for agents to get information from the environment and take action. Details are provided in Appendix \ref{App:C}. 
\subsubsection{Benchmarks and Metrics} 
\textbf{Multi-agent cooperative tasks.} We use VillagerBench~\cite{dong2024villageragent}, which includes three types of multi-agent tasks: construction cooperation, farm-to-table cooking and escape room challenge. Construction cooperation requires agents to understand the specified blueprint and complete the construction. Farm-to-table cooking requires the agent to perform hunting or harvesting according to the recipe to obtain ingredients and then gather all the ingredients together to craft the task object. The escape room challenge requires agents to individually activate the correct mechanisms to collectively solve the puzzle. 

There are three different evaluation metrics to assess the effectiveness of the method for all these three tasks, i.e., completion rate (\textbf{CR}), efficiency (\textbf{E}) and balanced agent utilization score (\textbf{BS}). Specifically, the completion rate represents task progress, with multiple indicators in each task for calculation. Assume the number of accessed indicators and total indicators amount are $I_{a}$ and $I_{t}$, completion rate can be defined as Equation \eqref{eq:7}:
\begin{equation}
  \label{eq:7}
  \textbf{CR} = \frac{I_{a}}{I_{t}} .
\end{equation}

Efficiency describes the ratio of the completion rate and the time agents take to execute the action. It measures the extent to which actions chosen by the agents contribute to the task completion. In an $N$ agents scenario, this metric can be calculated by the following equation, where $ET$ denotes the agents' execution time set, $t_j$ denotes the execution time of agent $j \in \{1,2,\dots, N\}$:
\begin{equation}
    \label{eq:8}
    \textbf{E} = \frac{\textbf{CR}}{\Sigma_jt_j}, t_j \in ET.
\end{equation}

BS aims to evaluate the distribution of workload. A higher BS should indicate better equilibrium, which means each agent’s active running time becomes more similar. Specifically:
\begin{equation}
  \label{eq:9}
  \textbf{BS} = 1-\frac{1}{N}\Sigma_j(std(\frac{|t_j -\min(ET)|}{T_{max}-\min(ET))}),
\end{equation}
where $T_{max}$ indicates the maximum action time for each task, actions exceeding this time limit will be considered as timed out. These evaluation metrics simultaneously assess the method's task completion ability and its capability to utilize multiple agents. Despite the three primary metrics \textbf{CR}, \textbf{E} and \textbf{BS}, some tasks incorporate auxiliary evaluation metrics such as view hit rate (\textbf{VHR}) and agent contribution rate (\textbf{ACR}). Detailed information on tasks and corresponding metric formulations are provided in Appendix \ref{App:A}.

\noindent\textbf{Single-agent tasks.} Following \citet{wangvoyager,wang2024jarvis,lioptimus}, we choose five representative items in Minecraft gameplay, i.e.,
Wood \includegraphics[scale=0.10,valign=c]{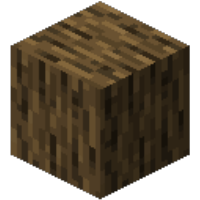}, Cobblestone \includegraphics[scale=0.10,valign=c]{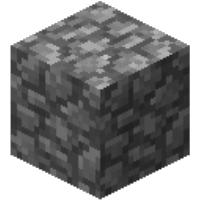}, Iron Ingot \includegraphics[scale=0.10,valign=c]{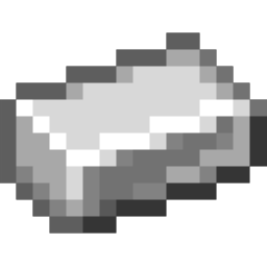}, Gold Ingot \includegraphics[scale=0.10,valign=c]{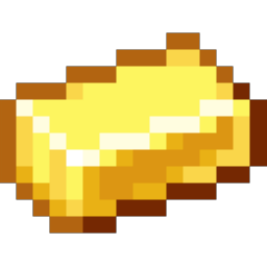} and Diamond \includegraphics[scale=0.10,valign=c]{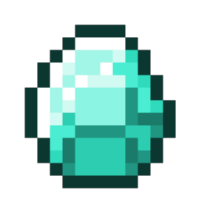}. The tasks require an agent to obtain each item from scratch with an empty inventory respectively, which is strongly relative to the agent's planning capability and long-term exploration ability. To evaluate methods on these tasks, we directly utilize success rate (\textbf{SR}) as the metric.

\subsection{Comparison with State-of-the-Arts}
\subsubsection{Main Results}
In this section, we compare our method with previous works. For multi-agent collaborative tasks, we compare with AgentVerse~\cite{chen2023agentverse} and VillagerAgent~\cite{dong2024villageragent}. For single-agent tasks, we compare with traditional reinforcement learning methods such as VPT~\cite{baker2022video}, STG-T~\cite{zhou2024learning}, PTGM~\cite{yuan2024pre}), and LLM-based methods such as vanilla ReAct~\cite{yaoreact}, Inner Mono~\cite{huang2023inner}, DEPS~\cite{wang2024describe}, Jarvis-1~\cite{wang2024jarvis}, Optimus-1~\cite{lioptimus} and Voyager~\cite{wangvoyager}. We use GPT-4o with default hyper-parameters for LLM-based methods.

\begin{table}[!t]
\centering
\begin{adjustbox}{width=\linewidth}
\small
\setlength{\tabcolsep}{6pt} 
\begin{tabular}{lccccc} 
\toprule
\multirow{2}{*}{\textbf{Method}} &  \multirow{2}{*}{\textbf{\makecell{No.\\\includegraphics[scale=0.04,valign=c]{steve.png}}}}& \multicolumn{1}{c}{\textbf{Construction}} & \multicolumn{1}{c}{\textbf{Escape}}&  \multicolumn{1}{c}{\textbf{Cooking}} &  \textbf{Avg.$^*$} \\
& & (\%) & (\%) & (\%) & (\%)\\
\midrule
AgentVerse & 2 & - & - & 91.41 & 91.41\\ 
\midrule
\multirow{2}{*}{{VillagerAgent}} & 2 & 91.09 & 92.15 & \textbf{95.45} & 92.00\\ 
& 3 & 87.51 & 75.99 & 63.36 & 79.93\\ 
\midrule
& 2 & \textbf{91.17} & 94.08 & 94.62 & 92.82\\
\rowcolor{black!15} \cellcolor{white}CausalMACE (Ours)& \cellcolor{white}3 & 90.50 & \textbf{96.34} & 92.41 & \textbf{93.32}\\
& 6 & 86.07 & 91.38 & 90.81 & 88.93\\
\bottomrule

\end{tabular}
\end{adjustbox}
\caption{\textbf{BS on each task.} \textit{No.} denotes the agent number and \textit{Avg.$^*$} is the weighted average based on the number of tasks. A higher BS represents a more balanced workload. 
}
\label{br}
\vspace{-0.5em}
\end{table}

\begin{table}[t]
\centering
\begin{adjustbox}{width=1\linewidth}
\begin{tabular}{lcccccc}
\toprule
\multirow{2}{*}{\textbf{Method}}   &  \textbf{Wood} &  \textbf{Cobblestone} &  \textbf{Iron}&  \textbf{Gold} &   \textbf{Diamond}  \\ 
&\includegraphics[scale=0.20,valign=c]{logos/wood.pdf}&\includegraphics[scale=0.20,valign=c]{logos/cobblestone.pdf}&\includegraphics[scale=0.15,valign=c]{logos/iron_ingot.pdf}&\includegraphics[scale=0.15,valign=c]{logos/gold_ingot.pdf}&\includegraphics[scale=0.20,valign=c]{logos/diamond.pdf}\\
\midrule
\multicolumn{6}{l}{\textbf{RL-based Agents}} \\
STG-T (NeurIPS'24)  & 6.0            & 0.0      & 0.0       & 0.0        & 0.0             \\
VPT-bc (NeurIPS'22)           & 20.0           & 18.0   & 0.0       &  0.0       &  0.0           \\
PTGM (ICLR'24)             & 63.0           & 59.0   & 0.0       & 0.0        & 0.0             \\
VPT-rl (NeurIPS'22)          & 99.0           & 99.0   & 60.0    &   0.0      & 15.0         \\           
\midrule
\multicolumn{6}{l}{\textbf{LLM-based Agents}} \\
ReAct (ICLR'23)          &  26.7           &  20.0  &   0.0    &    0.0      &   0.0          \\
Inner Mono (PMLR'23)   & 36.7           &  66.7  &   0.0    &    0.0      &   0.0          \\
DEPS (NeurIPS'24)         &  77.0           &  48.5  &  16.3  &  0.0        &   0.6        \\
Jarvis-1 (TPAMI'24)     &  91.6           &  94.2  &  33.8  &  14.5     &   9.2        \\
Optimus-1 (NeurIPS'24)    &  98.6           &  92.4  &  46.7  &  8.5      &   11.6       \\  
Voyager$^*$ (TMLR'24)  &  \textbf{100.0}          &  \textbf{100.0} &  76.5  &  52.9     &   29.4       \\ \midrule
\rowcolor{black!15}\textbf{CausalMACE (Ours)} &  \textbf{100.0}          &  \textbf{100.0} &  \textbf{82.3}  &  \textbf{70.6}     &   \textbf{41.1}        \\ \bottomrule
\end{tabular}
\end{adjustbox}
\caption{\textbf{Success rate on single-agent tasks.} * denotes that we re-conduct the method for adapting our settings.}
\label{tab:long-horizon}
\vspace{-0.75em}
\end{table}

The results on multi-agent collaborative tasks are listed in Table \ref{Main Result} and \ref{br}. Our method achieves superior performance over baseline approaches across most of the settings in both completion rate and efficiency, with particularly notable gains in construction cooperation. This result reveals the effectiveness of our method in multi-agent scenarios. While maintaining an excellent task completion rate and higher average BS, we observe a slight decrease in BS in several settings. These variations primarily stem from that some subtasks are naturally harder than others. When agents are assigned to a path with difficult subtasks, they take longer to finish. Such variations highlight broader challenges in workload balancing by estimating task durations, which are independent of our core method and need further research in the future. 

The results on single-agent collaborative tasks are listed in Table \ref{tab:long-horizon}. Our approach achieves perfect completion in basic single-agent scenarios while maintaining competitive performance as task complexity increases. This effectiveness is likely because, with our causal intervention, the agent can more reasonably schedule and execute subtasks. These results show that our method is also suitable for single-agent tasks.

\begin{figure*}[t] 
  \centering 
  \includegraphics[width=\textwidth]{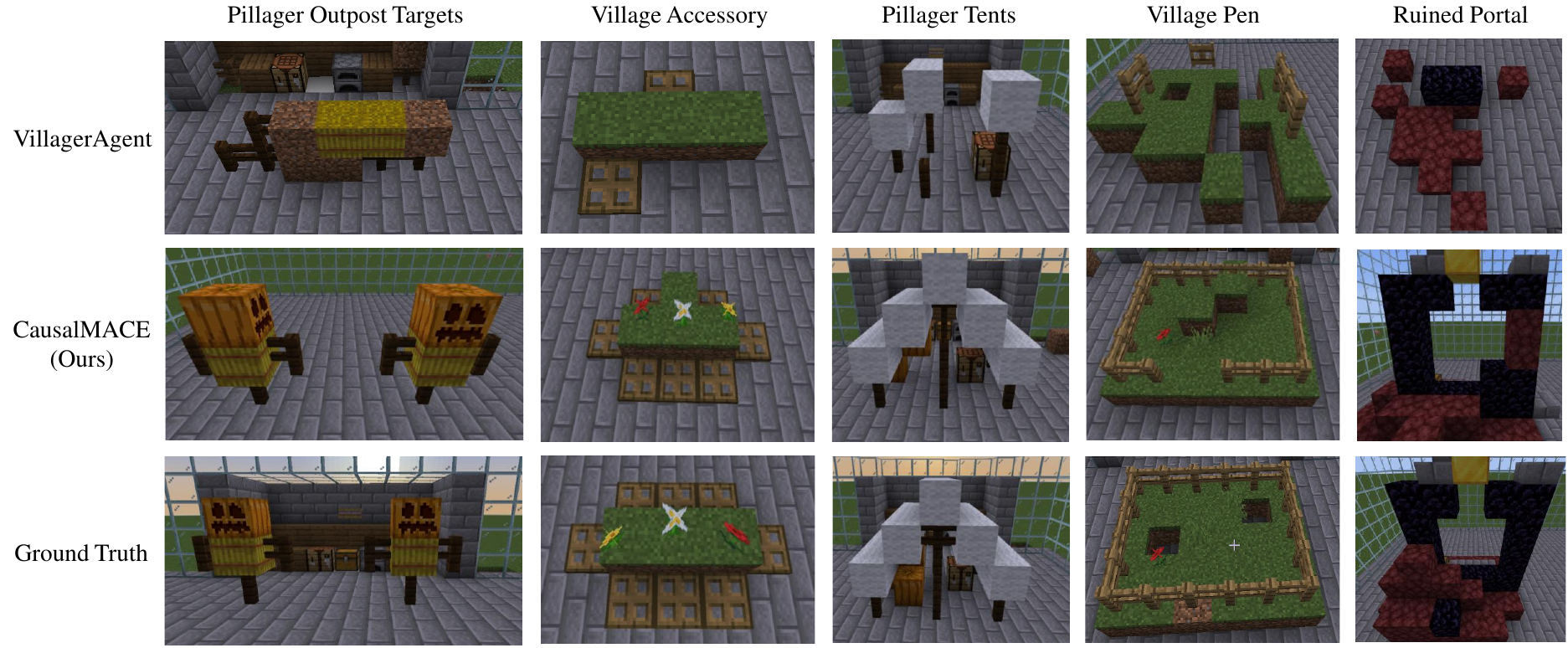} 
  \caption{\textbf{Visualization of construction cooperation.} VillagerAgent exhibits incoherent structural features, while CausalMACE maintains structural rationality and high consistency to ground truth with only minor local difference.} 
  \label{fig:casestudy} 
  \vspace{-0.75em}
\end{figure*}

\subsubsection{Dynamic Agent Numbers}
Here we discuss how agent quantity influences task performance. Existing study~\cite{dong2024villageragent} suggests that while scaling agent numbers initially improves outcomes, excessive agents may trigger resource competition and reduce effectiveness. However, we argue that for fixed-task scenarios, increasing agents could lower individual efficiency but still elevate collective performance, provided that coordination mechanisms and resource allocation are properly optimized. 

As demonstrated in Table \ref{Main Result}, the completion rate keeps increasing with the additional agents attending the task, except for the escape room challenge, where the difficulty of the task increases with the number of agents. 
These results indirectly reflect the adaptability of our framework in managing multi-agent collaboration under varying team sizes.

\subsection{Ablation Study}

\begin{table}[t]
\centering
\footnotesize
\begin{adjustbox}{width=1\linewidth}
\begin{tabular}{ccccccccc}
\toprule
\multirow{3.5}{*}{\textbf{Row}} & \multicolumn{3}{c}{\textbf{Setting}} &  \multicolumn{3}{c}{\textbf{Construction}}\\ 
\cmidrule(lr){2-4} \cmidrule(lr){5-7} & \multirow{2}{*}{\textbf{Busy Rate}} & \multirow{2}{*}{\textbf{Causal}} & \multirow{2}{*}{\textbf{Graph}} &
\textbf{CR} & \textbf{VHR} &  \textbf{Efficiency}   \ \\
&&&&  (\%)& (\%)&(\%/min)  \\
\midrule
1& \ding{56} & \ding{56} & \ding{56} & 52.68 & 55.99 & 3.93 \\
\midrule
2& \ding{51} & \ding{56} & \ding{56} & 57.53 & 56.70 & 5.31 \\
3& \ding{56} & \ding{56} & \ding{51} &57.92 &61.08&5.12 \\
\midrule
4& \ding{51} & \ding{56} & \ding{51} & 60.60 & 64.61 & 7.73 \\
5& \ding{56} & \ding{51} & \ding{51} & 72.49 & 75.20 & 6.98\\
\midrule
\rowcolor{black!15} \cellcolor{white}6& \ding{51} & \ding{51} &\ding{51} & \textbf{76.04} & \textbf{78.99} & \textbf{8.20} \\
\bottomrule
\end{tabular}
\end{adjustbox}
\caption{\textbf{Ablation study on construction cooperation.} We adopt 6 agents for each setting.}
\label{Ablation study}
\vspace{-1em}
\end{table}
We conduct ablation studies on construction tasks to validate our framework's key components. We specifically choose construction tasks due to their comprehensive requirements, including spatial planning, material collection, and real-time pathfinding in dynamic environments. The results are listed in Table \ref{Ablation study}.

\textbf{Row 6} in Table \ref{Ablation study} corresponds to the complete framework with all components included, which achieves the best completion rate and efficiency.

The removal of \textbf{Busy Rate} means to randomly assign paths to agents instead of workload-based allocation. \textbf{Row 5} shows it has few effects on the completion rate but has a significant impact on efficiency, as it may cause repetitive actions and result in an unbalanced workload among agents, increasing execution time. 

The removal of \textbf{Causal Intervention} indicates to utilize the initial graph $G_{init}$ without causal refinement. This leads to a significant decrease in the completion rate as shown in \textbf{Row 4}, representing the importance of causality to maintain the correct dependencies among subtasks.

The removal of \textbf{Graph} means to construct no graph and directly assigns subtasks to agents. Note that without \textbf{graph}, \textbf{causal intervention} is disabled automatically. \textbf{Row 2} demonstrates that the graph removal further influences the completion rate and efficiency simultaneously, indicating the foundational importance of the graph. 

\textbf{Row 1} and \textbf{Row 3} represent the joint ablation of Busy Rate with Causal Intervention and Graph, respectively. The results indicate that, regardless of whether Causal Intervention or Graph is present, the absence of Busy Rate leads to a degradation in efficiency.

\subsection{Case Study}
Figure \ref{fig:casestudy} compares the results of construction tasks between our method and VillagerAgent. While VillagerAgent produces almost no discernible structure, our method closely resembles the ground truth, with only a few blocks being incorrect or missing.

\begin{table}[t]
\centering
\footnotesize
\begin{tabular}{l|c|c}
\toprule
\multirow{2}{*}{\textbf{Method}} &Pillager Tents &Ruined Portal \\
&\includegraphics[scale=0.15,valign=c]{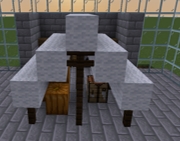} &  \includegraphics[scale=0.15,valign=c]{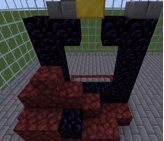}\\
\midrule
Luban &172.7  &311.2   \\
VillagerAgent &101.1  &175.7 \\
\rowcolor{black!15}\textbf{CausalMACE} &\textbf{81.0}  &\textbf{146.4}  \\
\bottomrule
\end{tabular}
\caption{\textbf{Average time consumption (seconds).} The image denotes the goal of construction. We adopt 6 agents for each method.}
\label{time}
\vspace{-0.75em}
\end{table}

We also compare CausalMACE with Luban \citep{guo2024luban}, which specializes in construction tasks. While Luban has demonstrated superior performance in pure construction, it does not address a complete task pipeline that includes resource acquisition and logistics management.
Therefore, we omit these phases for Luban in our comparison. Even with these advantages, Luban ultimately completes tasks with significantly higher time consumption. Table \ref{time} shows the time cost of different methods. Luban far exceeds the time consumption of other methods, highlighting CausalMACE's operational efficiency.

\section{Conclusion}
We present a novel global causality planning framework that significantly enhances the capabilities of multi-agent systems in open-world environments. By leveraging causality to manage and construct dependencies among subtasks from a global view, our approach ensures more efficient task arrangements. This addresses the limitations of existing multi-agent methods and enhances the overall effectiveness of task execution. Additionally, our findings suggest that incorporating causality into task planning can provide a more structured and efficient approach to managing complex tasks, leading to the possibility for further exploration.

\section*{Limitations}

While our framework demonstrates significant performance improvements across various Minecraft tasks, constraints still exist. The current causal intervention highly depends on the reasoning capacity of LLMs, making smaller models less viable options. This suggests that future improvements may leverage causality during model pretraining or fine-tuning for better adaptability.

\section*{Acknowledgments}
This research is supported by the National Natural Science Foundation of China (No. 62406267), Tencent Rhino-Bird Focused Research Program and the Guangzhou Municipal Science and Technology Project (No. 2025A04J4070).

\bibliography{main}
\clearpage
\appendix

\section{Action Functions}
\label{App:C}
We provide the action functions that are available for agents here.\\
\noindent\textbf{NavigateTo}(\textit{<pos>}). Move to the given position. The function returns True if the movement succeeds and False if the position is invalid or no valid path exists to the destination.\\
\noindent\textbf{CheckContainer}(\textit{container}). Access the specified container. The function returns information about the contents of the container.\\
\noindent\textbf{WithdrawItem}(\textit{container}, \textit{item}, \textit{amount}). Attempt to retrieve \textit{amount} of \textit{item} from the \textit{container}. The function returns the success status of the action. \\
\noindent\textbf{ScanEntities}(\textit{item}, \textit{distance}). Attempt to locate \textit{item} within \textit{distance}. The function returns the position of the item. If there is no \textit{item} within \textit{distance}, the function returns \textit{None}.\\
\noindent\textbf{Equip}(\textit{item}). Put \textit{item} in the main hand slot for usage or placement. The function returns True if it succeeds and False if the agent has no \textit{item} in inventory.\\
\noindent\textbf{PlaceBlock}(\textit{item}, \textit{<pos>}). Attempt to place \textit{item} at \textit{<pos>}. The function returns True if the placement succeeds, and False with the reason if the placement fails. The reasons include: the agent has not equipped the \textit{ item}, the agent is too far from \textit{<pos>}, the \textit{<pos>} is in the air without support blocks, the \textit{<pos>} is taken by the other block, etc.\\
\noindent\textbf{Handover}(\textit{item}, \textit{agent}). Give \textit{item} to \textit{agent}. The function returns the success status of the action. \\
\noindent\textbf{Craft}(\textit{item}, \textit{amount}). Attempt to craft \textit{amount} of \textit{item}. The function returns True if the crafting succeeds and False with the recipe and reason. Possible reasons include: the agent does not have enough ingredients in the inventory, the agent is too far from the crafting table, etc. \\
\noindent\textbf{Smelt}(\textit{item}, \textit{amount}, \textit{fuel}). Attempt to use the furnace to smelt \textit{item} with \textit{fuel}. The function returns the same information with \textbf{Craft}.\\
\noindent\textbf{MineBlock}(\textit{<pos>}). Attempt to dig the block at \textit{<pos>}. The function returns True if the mining succeeds and False with the reason. Possible reasons include: \textit{<pos>} is air, the agent is too far from \textit{<pos>}, the agent does not have proper tools, etc.\\
\noindent\textbf{Toggle}(\textit{<pos>}). Attempt to operate an interactive object at \textit{<pos>} (e.g., doors, pressure plates, buttons). The function returns the previous and current status of the object if the operation succeeds.\\

\noindent\textbf{UseOn}(\textit{target}). Attempt to use the tool in the main hand slot to \textit{target}. The function returns True if the action is valid and succeeds. Valid actions include: use shears on sheep, use bucket on cow, etc.

\noindent\textbf{Attack}(\textit{target}). Attack the target with the tool in the main hand slot. The function returns True if this attack succeeds.

\section{Reference Prompts}
\label{App:B}
We have shared the reference prompts we use for different modules below. Table \ref{prompt task decomposing} shows the prompt structure for task decomposition, while Table \ref{prompt dependencies} outlines the prompts for dependency prediction. Environmental perception and historical action summarization prompts are detailed in Table \ref{summary prompt}. The execution logic for agents is guided by Table \ref{agent execution prompt}, and the prompts for reflection are shown in Table \ref{prompt reflection}. Our demo video can be found in the attached file.

\begin{table*}
\centering
\resizebox{0.98\textwidth}{!}
{
\begin{tabular}{p{0.95\textwidth}}
\toprule
\large\textbf{Task Decomposition:}  \\
\hangindent=1em \hangafter=1 \qquad \small\texttt{Your current mission is to lead all the players and execute a set of specified tasks within the Minecraft environment.}\\
\hangindent=1em \hangafter=1 \qquad \small\texttt{-{}-{}- Background Information -{}-{}-}\\
\hangindent=1em \hangafter=1 \qquad \small\texttt{Our system manages the task as a Graph. In this turn, you need to decompose the tasks into steps, each step can be seen as a node. Next, we will construct the nodes into a graph. You can use the function to present each step, these functions should follow the rules below:}\\
\hangindent=1em \hangafter=1 \qquad \small\texttt{- Functions take some arguments, you need to decide how to design the function.}\\
\hangindent=1em \hangafter=1 \qquad \small\texttt{- Functions should be composed by atom steps, Atom steps are as follows: $\{available \: actions\}$.}\\
\hangindent=1em \hangafter=1 \qquad \small\texttt{- Functions should be executable, you need to make sure no matter which state the agent is in, it can achieve the goal by calling the function.}\\
\hangindent=1em \hangafter=1 \qquad \small\texttt{The function has the following JSON component:}\\
    
\hangindent=1em \hangafter=1 \qquad \small\texttt{\{}\\
\hangindent=1em \hangafter=1 \qquad \small\texttt{\quad "name": string, function name.}\\
\hangindent=1em \hangafter=1 \qquad \small\texttt{\quad "call": list of string, the argument list you input if you want to call the function.}\\
\hangindent=1em \hangafter=1 \qquad \small\texttt{\quad "description": description of the function.}\\
\hangindent=1em \hangafter=1 \qquad \small\texttt{\quad "function body": how does the function work.}\\
\hangindent=1em \hangafter=1 \qquad \small\texttt{\}}\\
\hangindent=1em \hangafter=1 \qquad \small\texttt{A node has the following JSON component:}\\
\hangindent=1em \hangafter=1 \qquad \small\texttt{\{}\\
\hangindent=1em \hangafter=1 \qquad \small\texttt{\quad "id": int, id of the step start from 1.}\\
\hangindent=1em \hangafter=1 \qquad \small\texttt{\quad "description": string, description of the step, more detail than a name, for example, place block needs position and facing, craft or collect items needs the number of items.}\\
\hangindent=1em \hangafter=1 \qquad \small\texttt{\quad "step": the function you choose and the arguments you need to input.}\\
\hangindent=1em \hangafter=1 \qquad \small\texttt{\}}\\
\hangindent=1em \hangafter=1 \qquad \small\texttt{*** Important Notice ***}\\
\hangindent=1em \hangafter=1 \qquad \small\texttt{- Task Decomposition: These atom steps should be small, specific, and executable with MineFlayer code, as you will be using MineFlayer to play Minecraft. Each atom step should contribute to the completion of the overall task. When necessary, the sub-tasks can be identical for faster task accomplishment. Be specific for the atom steps, for example, make sure to specify the materials needed.}\\
\hangindent=1em \hangafter=1 \qquad \small\texttt{- After all the steps are done, you need to make sure the whole task is done.}\\
\hangindent=1em \hangafter=1 \qquad \small\texttt{- In Minecraft, an item can be put in the agent's inventory, chest, or on the ground. You can use the item in the agent's inventory or chest, but you can not use the item on the ground.}\\
\hangindent=1em \hangafter=1 \qquad \small\texttt{- Integration and Finalization: In some tasks, you will need to integrate your individual efforts. For example, when crafting complicated stuff that requires various materials, after collecting them, you need to consolidate all the materials with one of the players.}\\
\hangindent=1em \hangafter=1 \qquad \small\texttt{Here is the query:}\\
\hangindent=1em \hangafter=1 \qquad \small\texttt{The environment information around: $\{env\}$.}\\
\hangindent=1em \hangafter=1 \qquad \small\texttt{The high-level task: $\{task\}$.}\\
\hangindent=1em \hangafter=1 \qquad \small\texttt{Your response should exclusively include:}\\
\hangindent=1em \hangafter=1 \qquad \small\texttt{- All Functions;}\\
\hangindent=1em \hangafter=1 \qquad \small\texttt{- Complete List of Nodes.}\\
\hangindent=1em \hangafter=1 \qquad \small\texttt{Formatted as: {"functions": [], "nodes": []}.}\\
\hangindent=1em \hangafter=1 \qquad \small\texttt{Your Response should contain ALL definitions of the functions and a COMPLETE list of nodes structed as ONE JSON.}\\

\bottomrule
\end{tabular}
}

\caption{Prompts for construction task decomposing.}
\label{prompt task decomposing}
\end{table*}

\begin{table*}
\centering
\resizebox{0.98\textwidth}{!}
{
\begin{tabular}{p{0.95\textwidth}}
\toprule
\large\textbf{Subtask Dependencies Prediction}  \\

\hangindent=1em \hangafter=1 \qquad \small\texttt{Your current mission is to lead all the players and execute a set of specified tasks within the Minecraft environment.}\\
\hangindent=1em \hangafter=1 \qquad \small\texttt{--- Background Information ---}\\
\hangindent=1em \hangafter=1 \qquad \small\texttt{Our system manages the task as a Hierarchical Causal Graph.
In this turn, I will give you a list of nodes, and you need to estimate whether certain nodes have a causal effect on other nodes.}\\
\hangindent=1em \hangafter=1 \qquad \small\texttt{You should also check if other nodes affect the node. 
If node a have causal effect on node x and node y, which means if a is not completed, x and y can not start, you should add new jsons to the edge list \{"chosen\_node": a, "target\_node": x\}, \{"chosen\_node": a, "target\_node": y\}.}\\
\hangindent=1em \hangafter=1 \qquad \small\texttt{If node z has a causal effect on the certain node a, which means if z is not completed, a can not start, you should add a new JSON to the edge list \{"chosen\_node": z, "target\_node": a\}.}\\
\hangindent=1em \hangafter=1 \qquad \small\texttt{*** Important Notice ***}\\
\hangindent=1em \hangafter=1 \qquad \small\texttt{The causal effect should follow some basic Minecraft rules.}\\
\hangindent=1em \hangafter=1 \qquad \small\texttt{- You can not place/use something unless you get it from container first;}\\
\hangindent=1em \hangafter=1 \qquad \small\texttt{- You can not place something if you have not equipped it;}\\
\hangindent=1em \hangafter=1 \qquad \small\texttt{- The block at the lower place should be placed first, and the block at the higher place should be placed later.}\\
\hangindent=1em \hangafter=1 \qquad \small\texttt{Here is the query: $\{nodes\}$.}\\
\hangindent=1em \hangafter=1 \qquad \small\texttt{The node you need to deal with now: $\{node \: id\}$.}\\
\hangindent=1em \hangafter=1 \qquad \small\texttt{Your response should exclusively include:}\\
\hangindent=1em \hangafter=1 \qquad \small\texttt{- ALL Causal Effect of the chosen node.}\\
\hangindent=1em \hangafter=1 \qquad \small\texttt{Formatted as: \{"causal effect": []\}.}\\
\hangindent=1em \hangafter=1 \qquad \small\texttt{Think step by step. Your Response should formatted as ONE JSON.}\\

\bottomrule
\end{tabular}
}
\caption{Prompts for construction dependencies prediction.}
\label{prompt dependencies}
\end{table*}

\begin{table*}
\centering
\resizebox{0.98\textwidth}{!}
{
\begin{tabular}{p{0.95\textwidth}}
\toprule
\large\textbf{Environmental Data Summarization:}  \\
\hangindent=1em \hangafter=1 \qquad \small\texttt{You are a helpful assistant in Minecraft. Based on the environment info and the task, extract the key information and summarize the environment info in a concise and informative way. You should focus on the entities, blocks and creatures in the environment, and provide a summary of the environment info.}\\
\hangindent=1em \hangafter=1 \qquad \small\texttt{The environment info:
$\{info\}$}\\
\hangindent=1em \hangafter=1 \qquad \small\texttt{The task: $\{task\}$.}\\
\hangindent=1em \hangafter=1 \qquad \small\texttt{Return with Entity, Blocks, Creatures, Interactive-Items and give all these positions of these blocks and entities like chest, crafting table, furnace, animals and plants.}\\
\midrule
\large\textbf{History Action Summarization:}  \\
\hangindent=1em \hangafter=1 \qquad \small\texttt{You are a helpful assistant in Minecraft. Your name is $\{name\}$. Your task is to create a concise running summary of actions and information results in the provided text, focusing on key and potentially important information to remember.}\\

\hangindent=1em \hangafter=1 \qquad \small\texttt{You will receive the current summary and your latest actions. Combine them, adding relevant key information from the latest development in 1st person's past tense and keeping the summary concise. The subject of the sentence should be $\{name\}$.}\\

\hangindent=1em \hangafter=1 \qquad \small\texttt{Summary So Far: $\{summary\}$.}\\

\hangindent=1em \hangafter=1 \qquad \small\texttt{Latest Actions: $\{actions\}$.}

\hangindent=1em \hangafter=1 \qquad \small\texttt{Return with your summary.}\\

\bottomrule
\end{tabular}
}
\caption{Prompts for data and history summarizing.}
\label{summary prompt}
\end{table*}

\begin{table*}
\centering
\resizebox{0.98\textwidth}{!}
{
\begin{tabular}{p{0.95\textwidth}}
\toprule
\large\textbf{Agent Execution}  \\

\hangindent=1em \hangafter=1 \qquad \small\texttt{The relevant data of task: $\{task\}$.}\\
\hangindent=1em \hangafter=1 \qquad \small\texttt{The state of the agent: $\{state\}$.}\\
\hangindent=1em \hangafter=1 \qquad \small\texttt{The agent's actions in the last time segment partially: $\{history\}$.}\\
\hangindent=1em \hangafter=1 \qquad \small\texttt{The environment info: $\{env\}$.}\\
\hangindent=1em \hangafter=1 \qquad \small\texttt{The agent's inventory status: $\{inventory\}$.}\\
\hangindent=1em \hangafter=1 \qquad \small\texttt{The Minecraft knowledge card:}\\
\hangindent=1em \hangafter=1 \qquad \small\texttt{\quad - In minecraft world x,z is the horizontal coordinate, y is the vertical coordinate.}\\
\hangindent=1em \hangafter=1 \qquad \small\texttt{\quad - You can place the block to the world.}\\
\hangindent=1em \hangafter=1 \qquad \small\texttt{\quad - You can find the item in the chest. Item in the chest can not directly be seen or used, take it out and use it or equip it.}\\
\hangindent=1em \hangafter=1 \qquad \small\texttt{\quad - If there is no item in the chest, maybe you can find the item at the other chest, get it from the other agent, dig it up or craft it.}\\
\hangindent=1em \hangafter=1 \qquad \small\texttt{\quad - One bucket can hold one item, if you want to get more items, you need to get more buckets at first.}\\
\hangindent=1em \hangafter=1 \qquad \small\texttt{\quad - Do not change the blocks other agents placed without permission.}\\
\hangindent=1em \hangafter=1 \qquad \small\texttt{\quad - Your height is two block, if you need to move to somewhere and you find your target position is in the air, You can check a location that is one level below your target location.}\\
\hangindent=1em \hangafter=1 \qquad \small\texttt{\quad - You should first try to take the action, only rethink if fail to do so.}\\
\hangindent=1em \hangafter=1 \qquad \small\texttt{\quad - If you think you have something in your inventory, you should check your inventory instead of scanning around.}\\
\hangindent=1em \hangafter=1 \qquad \small\texttt{\quad - Do not ask other players to give you something, because they need to do their work. Only require item if you fails to get it from env or chest.}\\
\hangindent=1em \hangafter=1 \qquad \small\texttt{\quad - If you need to place a certain block at a certain place, and the place is already occupied by the kind of block you need to place, you should treat it as you have completed the task. Note: only do this if you can make sure that the block there is the same block as the block you need to place!}\\
\hangindent=1em \hangafter=1 \qquad \small\texttt{Think by steps and take action.}\\

\bottomrule
\end{tabular}
}
\caption{Prompts for agent execution.}
\label{agent execution prompt}
\end{table*}

\begin{table*}
\centering
\resizebox{0.98\textwidth}{!}
{
\begin{tabular}{p{0.95\textwidth}}
\toprule
\large\textbf{Agent Reflection}  \\

\hangindent=1em \hangafter=1 \qquad \small\texttt{You are in a Minecraft world. You are an agent. You need to use the action history compared with the task description to check whether the task is completed.}\\
\hangindent=1em \hangafter=1 \qquad \small\texttt{The check-structure:}\\
\hangindent=1em \hangafter=1 \qquad \small\texttt{\{}\\
\hangindent=1em \hangafter=1 \qquad \small\texttt{\quad "reasoning": str, the reasoning process.}\\
\hangindent=1em \hangafter=1 \qquad \small\texttt{\quad "summary": str, the summary of the vital information of action history with detailed position number and other parameters, which are not included in the task description.}\\
\hangindent=1em \hangafter=1 \qquad \small\texttt{\quad "status": bool, whether the task is completed.}\\
\hangindent=1em \hangafter=1 \qquad \small\texttt{\}}\\
\hangindent=1em \hangafter=1 \qquad \small\texttt{Now you have tried to complete the task.}\\
\hangindent=1em \hangafter=1 \qquad \small\texttt{The task description is: $\{task\}$.}\\
\hangindent=1em \hangafter=1 \qquad \small\texttt{The action history is: $\{history\}$.}\\
\hangindent=1em \hangafter=1 \qquad \small\texttt{The state of the agent is: $\{state\}$.}\\
\hangindent=1em \hangafter=1 \qquad \small\texttt{Please check whether the task is completed and return a check-structure JSON.}\\

\bottomrule
\end{tabular}
}
\caption{Prompts for agent reflection.}
\label{prompt reflection}
\end{table*}

\section{Detailed Experimental Setups and Metrics}
\label{App:A}
This section will describe our experimental setups and metrics in more detail. For multi-agent cooperative tasks, we conduct experiments on three different scenarios, including construction cooperation, farm-to-table cooking and escape room challenge. All the settings (e.g., task amounts, metrics) follow VillagerBench~\cite{dong2024villageragent}. For single-agent tasks, we conduct experiments on item gathering following~\cite{wang2024jarvis,lioptimus}. We will introduce the settings respectively.
\subsection{Construction Cooperation}
Construction cooperation requires agents to construct a structure or a building following a specified blueprint which includes various blocks. Since the process of obtaining blocks can be complex, considering the inherent task difficulty, the requirement for agents to mine blocks is omitted, which keeps the setting consistent with VillagerBench~\cite{dong2024villageragent}. This task evaluates the understanding of spatial dependencies and task requirements in multi-agent collaboration. 

\textbf{Completion Rate in Construction Cooperation}: The completion rate in this task represents the block hit rate, which is the ratio of correctly placed blocks to the total number of blocks. Only blocks with the correct orientation, type and position are considered correctly placed. The formal definition is as follows:
\begin{equation}
    \textbf{CR} = \frac{\operatorname{card}(CP)}{\operatorname{card}(B)},
\label{eq:10}
\end{equation}
where $B$ indicates the blocks set of the blueprint, $CP \subseteq B$ indicates the set of correctly placed blocks, $\operatorname{card}(\cdot)$ indicates the cardinality of a set. A higher completion rate indicates a closer match to the blueprint, reflecting the agents’ ability to accurately execute the construction plan.

\textbf{View Hit Rate in Construction Cooperation}: View hit rate (\textbf{VHR}) is a special metric for construction cooperation, it is used to assess the structural integrity and visual coherence of a construction from multiple viewpoints. By comparing the overlap between the constructed structure and the expected structure across a predefined set of viewpoints, a higher \textbf{VHR} score indicates a closer match between the actual and expected structures, suggesting better structural integrity and visual coherence. This metric is calculated by the Intersection over Union (IoU) of the constructed structure with the blueprint across various viewpoints. The formal definition is as follows:
\begin{equation}
    \textbf{VHR} = \frac{1}{V}\sum_{i=1}^V{IoU(\frac{P_i}{B_i})},
\label{eq:11}
\end{equation}
where $P_i$ indicates the constructed structure seen from viewpoint $i$, $B_i$ indicates what should the structure looks like from viewpoint $i$. $V$ is the number of viewpoints.

\subsection{Farm-to-Table Cooking}
Farm-to-table cooking requires agents to prepare dishes according to the given recipe. In this task, agents need to gather information from the environment and adjust their strategies to collect ingredients from containers or sometimes to harvest crops and hunt animals in the wild. This task evaluates the agents' environmental perception and coordination capacities.

\textbf{Completion Rate in Farm-to-Table Cooking}: The completion rate in farm-to-table cooking represents the progress of the recipe. Assume that the recipe for preparing a certain dish needs $M$ ingredients and $N$ actions, the completion rate should be:
\begin{equation}
    \textbf{CR} = \frac{\sum_{i=1}^Mk_{mi}\times S_{mi} + \sum_{j=1}^Nk_{aj}\times S_{aj}}{\sum_{i=1}^MS_{mi} + \sum_{j=1}^N S_{aj}},
\end{equation}
where $k$ represents the status of a certain ingredient or action. If the ingredient is gathered or the action is taken successfully, $k$ should be one, otherwise, it should be zero. $S_{mi}$ represents the score of $i$-th ingredient and $S_{aj}$ represents the score of $j$-th action. The higher completion rate represents closer to finishing preparing the dish.

\textbf{Agent Contribution Rate}: Agent contribution rate (\textbf{ACR}) measures if all agents contribute to the task. Assume the score contributed by agent $i$ is $C_i$, average score contributed by each agent is $C_{avg}$, we first calculate the standard deviation $\sigma(C)$ of contribution among all agents:
\begin{equation}
    \sigma(C) =  \sqrt{\frac{1}{N} \sum_{i=1}^{N} (C_i - C_{avg})^2},
\end{equation}
where $N$ refers to the number of agents. The deviation $\sigma(C)$ reflects the degree of dispersion of the contribution, which means a smaller $\sigma(C)$ leads to a more balanced distribution. We then calculate the agent contribution rate by normalizing $\sigma(C)$:
\begin{equation}
    \textbf{ACR} = 1 - \frac{\sigma(C)-\sigma_{min}}{\sigma_{max}-\sigma_{min}},
\end{equation}
where $\sigma_{min}$ refers to the minimum possible deviation value, i.e., all agents contribute the same score, $\sigma_{max}$ refers to the maximum possible deviation value, i.e., one agent finishes all the process. 

\subsection{Escape Room Challenge}
The escape room challenge tests the agents’ ability to work together for synchronization and sequential execution. In this task, agents will need to activate certain objects in the environment in order or simultaneously. 

\textbf{Completion Rate in Escape Room Challenge}
completion rate in the escape room challenge represents the finishing rate of room tasks. The formal definition is as follows:
\begin{equation}
    \textbf{CR} = \frac{\sum_{i=1}^N\frac{c_i}{a_i}\times S_i}{\sum_{i=1}^N S_i},
\end{equation}
where $N$ denotes total number of the rooms, $i$ denotes $i$-th room, $c_i$ denotes the number of correct conditions, $a_i$ denotes all the conditions, $S_i$ denotes the assigned score of each task.

\subsection{Item Gathering}
Item gathering challenges a single agent to collect specific items in the open-world environment from scratch. The agent must navigate through different areas, identify the required items and retrieve them while managing potential obstacles. The task requires the agent to develop efficient strategies for exploration, item identification and collection. This task evaluates the agent's ability to autonomously explore the environment, make decisions based on available resources and execute the task while overcoming spatial and temporal challenges.

Specifically, the first task is to collect one log \includegraphics[scale=0.10,valign=c]{logos/wood.pdf}. This is the most basic task in Minecraft because it serves as the foundational step for many other actions within the game. A log is one of the first resources that players encounter and collecting it unlocks a variety of crafting possibilities, such as turning it into planks, crafting tables, or other essential tools. The next task is to get a cobblestone \includegraphics[scale=0.10,valign=c]{logos/cobblestone.pdf}. This task is the most basic mining task. It tests whether the agent has the ability to craft and the necessary proficiency in chopping trees, as mining stone requires a wooden pickaxe as a tool, and this tool must be crafted by the agent itself. 

If the agent is able to complete the two basic tasks mentioned above, it demonstrates the capability to tackle more advanced tasks. A typical advanced task is to obtain an iron ingot \includegraphics[scale=0.10,valign=c]{logos/iron_ingot.pdf}, which requires mining and using a furnace to smelt ores. Acquiring iron is a fundamental step toward obtaining more advanced resources. The ability to obtain iron ingots indicates that the agent not only has proficient mining skills but is also capable of performing a series of operations such as searching for ores and smelting them.

The last two tasks are obtain a gold ingot \includegraphics[scale=0.10,valign=c]{logos/gold_ingot.pdf} and a diamond \includegraphics[scale=0.10,valign=c]{logos/diamond.pdf}. The last two tasks are the most difficult in the item-gathering objectives. In Minecraft, gold and diamonds require an iron pickaxe to be obtained, which requires 3 iron ingots to be crafted. Therefore, if the agent is not skilled enough to complete all tasks up to acquiring iron ingots, they will not be able to complete these tasks. Even if the agent is skilled at obtaining iron and crafting an iron pickaxe, they will still need to explore underground sufficiently to ensure they find gold ore and diamond ore. Successfully completing these two tasks indicates that the agent is capable of performing high-level tasks in Minecraft.

For metrics on item gathering, we adopt the success rate. Assume that the agent makes $M$ attempts on a certain task, and $N$ of them succeed, the success rate (\textbf{SR}) should be:
\begin{equation}
    \textbf{SR} = \frac{N}{M}.
\end{equation}

A higher success rate indicates that the agent is more skilled at completing the task.

\end{document}